\documentclass[10pt,twocolumn,letterpaper]{article}

\usepackage{wacv} 

\usepackage{graphicx}
\usepackage{amsmath}
\usepackage{amssymb}
\usepackage{booktabs}

\usepackage{listings}
\usepackage{algorithm}
\usepackage{algpseudocode}
\algnewcommand\algorithmicforeach{\textbf{for each}}
\algdef{S}[FOR]{ForEach}[1]{\algorithmicforeach\ #1\ \algorithmicdo}
\usepackage{multirow}

\usepackage[pagebackref,breaklinks,colorlinks]{hyperref}

\usepackage[capitalize]{cleveref}
\crefname{section}{Sec.}{Secs.}
\Crefname{section}{Section}{Sections}
\Crefname{table}{Table}{Tables}
\crefname{table}{Tab.}{Tabs.}

\def\confName{WACV}
\def\confYear{2024}

\newtheorem{example}{Example}
\newtheorem{definition}{Definition}

\def\bi{\begin{itemize}}
	
	\def\ei{\end{itemize}}
\def\beq{\begin{equation}}
	\def\eeq#1{\label{#1}\end{equation}}
\def\ba{\begin{array}}
	\def\ea{\end{array}}

\long\def\BOC#1\EOC{\message{(Commented text )}}
\long\def\BOCC#1\EOCC{\message{(Commented text )}}
\long\def\BOCCC#1\EOCCC{\message{(Commented text )}}

\long\def\NBB#1{}

\def\ours{\text{CRCG}}

\usepackage[normalem]{ulem}

\begin{document}
	

\title{Think before You Simulate: Symbolic Reasoning to Orchestrate Neural Computation for Counterfactual Question Answering}
	
\author{Adam Ishay$^1$, Zhun Yang$^1$,  Joohyung Lee$^{1,2}$ \\
		$^1$ Arizona State University, AZ, USA\\
		{\tt\small \{aishay, zyang90, joolee\}@asu.edu}
	\and
	Ilgu Kang$^2$, Dongjae Lim$^2$  \\
	$^2$ Samsung Research, S. Korea\\
	{\tt\small \{ilgu.kang, dongjae.lim\}@samsung.com}
}
\maketitle

\begin{abstract}
Causal and temporal reasoning about video dynamics is a challenging problem. While neuro-symbolic models that combine symbolic reasoning with neural-based perception and prediction have shown promise, they exhibit  limitations, especially in answering counterfactual questions.  This paper introduces a method to enhance a neuro-symbolic model for counterfactual reasoning, leveraging symbolic reasoning about causal relations among events. We define the notion of a causal graph to represent such relations and use Answer Set Programming (ASP), a declarative logic programming method, to find how to coordinate perception and simulation modules. We validate the effectiveness of our approach on two benchmarks, CLEVRER and CRAFT. Our enhancement achieves state-of-the-art performance on the CLEVRER challenge, significantly outperforming existing models. In the case of the CRAFT benchmark, we leverage a large pre-trained language model, such as GPT-3.5 and GPT-4, as  a proxy for a dynamics simulator. Our findings show that this method can further improve its performance on counterfactual questions by providing alternative prompts instructed by symbolic causal reasoning. 
\end{abstract}

\vspace{-2mm}
\section{Introduction} \label{sec:intro}

The ability to recognize object movement and reason about its dynamics is a fundamental aspect of human cognition \cite{ullman17mind}. Deep neural networks have shown remarkable progress in recognizing patterns in complex visual and language inputs \cite{shrivastava17learning,huang21seeing,meng22adavit,song22learning}, but answering questions involving temporal and causal structures in video dynamics remains a significant challenge in AI. This is particularly true when dealing with hypothetical questions such as predictive and counterfactual ones \cite{wagner18answering,yi19clevrer,sampat21clevr_hyp}.
Highlighting this issue, Yi et al. \cite{yi19clevrer} introduced a challenging benchmark known as CLEVRER\footnote{http://clevrer.csail.mit.edu.}. This comprises four types of questions about videos featuring the movement of various objects, each differing in shape, color, and material. They observed that previous state-of-the-art end-to-end models for visual QA, such as TbD-Net \cite{mascharka18transparency}, MAC \cite{hudson2018compositional}, and IEP \cite{johnson2017inferring}, failed to perform well on the CLEVRER benchmark. In response, they proposed a neuro-symbolic hybrid AI model called NS-DR that outperforms these models. The key strategy involves integrating (i) neural components that recognize objects and events and simulate dynamics of objects with (ii) symbolic components that aggregate the outputs of the neural components and apply symbolic logic to answer the natural language questions. Further improvements were made along the same architecture, enhancing the neural components for more accurate perception and prediction. For instance, VRDP \cite{ding2021dynamic} incorporates a differentiable physics engine that infers explicit physical properties and uses this knowledge to yield better simulations. However, these models still have room for improvement, particularly concerning counterfactual questions.

On the other hand, a recent end-to-end neural model called Aloe \cite{ding21attention}, which is based on the self-attention mechanism, has demonstrated significant performance improvements when compared to the earlier neural models. This finding was further bolstered by ODDN-Aloe \cite{tang22object}, whose performance has been found to be on par with VRDP. However, it's important to note that these neural models still lack the transparency and interpretability that is often required.

In this paper, we argue that previous neuro-symbolic models have not fully utilized the strength of the neuro-symbolic approach. Instead of using symbolic reasoning only to aggregate information from neural components, we propose incorporating symbolic reasoning at the front as well to orchestrate between neural perception and neural simulation. 
Specifically, for counterfactual question answering, our method constructs a causal graph by observing the video and taking into account the objects that are intervened upon in the counterfactual question. We then compute the causal effects of this change and use the result to trigger simulation only when needed, starting from the relevant frames. This contrasts with the previous neuro-symbolic models which blindly apply simulation from the beginning. 
For the computation of a causal graph, we use Answer Set Programming (ASP) \cite{lif08,bre11}, a declarative logic programming method. 
We claim that our approach enhances baselines as long as perception is more accurate than simulation, which is usually the case for most baselines.  
Additionally, our model outperforms the baseline neuro-symbolic models even with the use of the same perception and simulation modules from them. 

We validate the effectiveness of our method by applying it to two benchmarks, CLEVRER and CRAFT \cite{ates20craft}. For the CLEVRER task, we also enhance answering the other types of questions by augmenting baseline models with additional modules, thanks to the modularity of the neuro-symbolic architecture. As a result, we achieve the state-of-the-art result on CLEVRER, outperforming all the models above.

We have also discovered an intriguing use case for large language models (LLMs).
Our method assumes the availability of a simulator, but we couldn't find a publicly available simulator for the CRAFT dataset. Instead of constructing a new simulator, we use an LLM, such as GPT-3.5 and GPT-4 \cite{brown2020language,openai23gpt4}, as a proxy simulator.
We provide GPT-x (x $\in\{3.5, 4\}$) with the natural language descriptions of the scenes in the dataset and use it to answer counterfactual questions.
Surprisingly, the vanilla GPT-x reasoning exhibits reasonable performance for the textual descriptions of visual scenes. Moreover, by applying our method, we can further enhance the performance by determining whether a counterfactual question can be answered using factual states, as demonstrated by the causal graph.
Since GPT-x handles factual questions more effectively than counterfactual questions, our method significantly improves the accuracy of GPT-x answers. Essentially, our method can be considered a new way of prompting GPT-x for counterfactual questions.

In summary, this paper first introduces a graphical model that formalizes causal and temporal relations among events. Second, we implement the model's computation in the declarative programming language ASP, which we use to improve counterfactual event prediction. Third, we demonstrate the effectiveness of our method by achieving  state-of-the-art performance on CLEVRER. Finally, we demonstrate the visual reasoning capability of an LLM and show how it can be further enhanced through our counterfactual reasoning approach.
		
The paper is organized as follows. 
Section~\ref{sec:prelim} provides a brief overview of the necessary background information. In Section~\ref{sec:math_model}, we introduce a graphical model that describes causal relationships among temporal events, and Section~\ref{sec:asp-impl} presents its implementation in ASP. Sections~\ref{sec:clevrer-experiment} and \ref{sec:craft-experiment} present experimental results with CLEVRER and CRAFT. 

The implementation of our method is publicly available  at 
\url{https://github.com/azreasoners/crcg}.
			
\section{Preliminaries}  \label{sec:prelim}
	
\subsection{Neuro-Symbolic Models and CLEVRER} \label{ssec:clevrer}

{\em NS-DR} explicitly combines perception, language, and physical dynamics through symbolic representation. It consists of the following modules. \\[-1em]

\begin{itemize}\setlength\itemsep{0em}
	\item  The question parser translates the question and answer options to a functional program, which is passed to the program executor.
	
	\item The video frame parser (perception module) identifies objects and their trajectories in the video. It returns 
	object trajectories and the intrinsic attributes of objects.
	
	\item The dynamics predictor (simulation module) learns the dynamics of objects  
	by training on object masks proposed by the video frame parser. 

	\item  The program executor is a symbolic reasoner that executes the functional program. It consists of several functional modules implemented in Python, which query the output of the dynamics predictor (such as getting the post-video or counterfactual collision events), or perform logical operations (such as recursively checking the ancestors of collision events). 	
\end{itemize}

A few enhancements were proposed building upon the structure of NS-DR. 
To avoid dense annotations for visual attributes and physical events, DCL \cite{chen21grounding} applies concept learning through weak supervision using question-answer pairs associated with videos. However, its accuracy is only marginally better than that of NS-DR. 

VRDP \cite{ding2021dynamic} has a similar architecture, but has a better dynamics predictor. It integrates a differentiable physics engine that infers explicit physical properties, such as mass and velocity, from object-centric representations and uses them to run simulations. Its overall performance on CLEVRER is slightly worse than that of ODDN-Aloe except for counterfactual questions. 

In all the models above, symbolic reasoning is applied at the end after aggregating all the information from other components to derive the answer.

\begin{figure*}[t!]
	\centering
\includegraphics[width=0.8\textwidth,height=3.2cm]{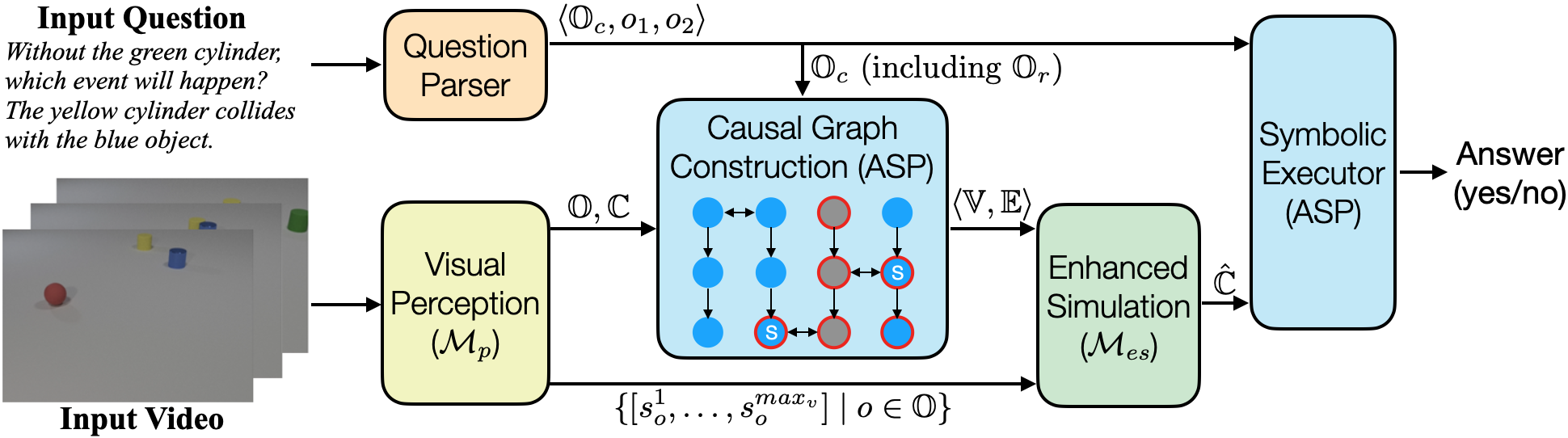}
	\caption{Overview of $\ours$ --- the enhanced counterfactual reasoning using a causal graph. The Causal Graph Construction and the Symbolic Executor modules are realized by ASP and perform symbolic computation.}
	\label{fig:our_model}
 \vspace{-2mm}
\end{figure*}

\subsection{Answer Set Programming} \label{ssec:asp}

Answer Set Programming (ASP) \cite{lif08,bre11} is a logic programming paradigm that allows for declarative reasoning in knowledge-intensive applications. It is based on the answer set semantics of logic programs \cite{gel88}, which enables the expression of causal reasoning, default reasoning, aggregates, and various other constraints. There are several efficient solvers, such as {\sc clingo}, {\sc dlv}, and {\sc wasp}.  

ASP has primarily been applied in symbolic domains, but there are some exceptions that apply ASP in conjunction with visual perception, as demonstrated in \cite{aditya15visual,suchan18visual,suchan2019out,khan2019visual,sautory21hyster}.
We use {\sc clingo} v5.3.0 as an ASP solver.
For the language of {\sc clingo}, we refer the reader to the textbook \cite{lifschitz19answer} or the {\sc clingo} manual.\footnote{\url{https://github.com/potassco/guide/releases/tag/v2.2.0}}

\section{Counterfactual Reasoning with Causal Graphs}\label{sec:math_model}

In videos depicting the objects moving and colliding with each other, the {\em state} of an object $o$ at a frame $t$ includes the information about its location, speed, direction,  mass, etc.
We aim to investigate the causal relationships between these states resulting from changes described in the counterfactual question. We will explore how identifying these causal relations can aid in counterfactual reasoning.

Consider a video ${\cal V}$ depicting a set $\mathbb{O}$ of objects. 
Let $Q=\langle \mathbb{O}_c, o_1, o_2 \rangle$ represent a counterfactual question ``if we intervene in (that is, remove or replace) some objects $\mathbb{O}_c$ ($\subseteq \mathbb{O}$), will there be a collision 
between objects $o_1$ and $o_2$?''
{Let $\mathbb{O}_r$ ($\subseteq \mathbb{O}_c$) denote the set of removed objects in $\mathbb{O}_c$ and let  $\mathbb{O}_u$ denote $\mathbb{O}\setminus\mathbb{O}_r$.}

Let ${\cal M}_p$ denote a {\em perception model} that detects  object states and collisions 
in the video:
\begin{align}\label{eq:m_p}
\{[s_o^1,\dots,s_o^{max_v}] \mid o\in \mathbb{O}\}, ~ \mathbb{C} \leftarrow {\cal M}_p({\cal V}),
\end{align}
where $s_o^t$ denotes the state of object $o$ at frame $t$ returned by the perception model (known as the {\em perception state}); 
$max_v$ is the total number of frames in a video; 
and $\mathbb{C}$ is a set of collision events of the form $\langle i,j,t \rangle$ found by the perception model, representing that objects $i$ and $j$ collide 
at frame $t$.

Let ${\cal M}_s$ denote a {\em simulation model} that takes the states of $\mathbb{O}_u$ at frame $t$ as input and outputs their states at frame $t+1$, and the 
{collisions} happened at frame $t$ in the simulation: 
\begin{align*}
\{\hat{s}_o^{t+1} \mid o\in \mathbb{O}_u\}, ~ \hat{\mathbb{C}}^t \leftarrow {\cal M}_s(\{\hat{s}_o^{t} \mid o\in \mathbb{O}_u\}) 
\end{align*}
where $\hat{s}_o^t$ denotes the state of object $o$ at frame $t$ returned by the simulation model ({\em simulated state}); 
$\hat{\mathbb{C}}^t$ is a set of 
{detected collisions at frame $t$;}
and the initial states are from perception, i.e., $\hat{s}_o^1 = s_o^1$ for $o\in \mathbb{O}_u$.

When comparing known perception states $s_o^t$ to simulated states $\hat{s}_o^t$, we typically find that the latter are less accurate, and that simulation error accumulates over frames.
To answer a counterfactual question, we aim to use the perception states for as long as possible, only switching to the simulated states once the perception states have been ``affected" by the intervened objects.


\subsection{A Causal Graph with Temporal Events} 

Our method, which we call $\ours$ (Counterfactual Reasoning using a Causal Graph), uses a pipeline shown in Figure~\ref{fig:our_model}. The pipeline begins with the Causal Graph Construction module that 
takes as input the intervened objects $\mathbb{O}_c$ from the Question Parser, and the objects $\mathbb{O}$ and in-video collisions $\mathbb{C}$ from the Visual Perception module.
The module constructs a graph $\langle \mathbb{V}, \mathbb{E} \rangle$ using the given inputs.

The Causal Graph Construction module first obtains $[t_1, \dots, t_k]$ from $\mathbb{C}$ as an ordered list of frames in the video when 
{collisions occur.}
Since only $k$ frames introduce causal relations, 
{we model the causal relations in video ${\cal V}$ with a {\em causal graph} $\langle \mathbb{V}, \mathbb{E} \rangle$ where}
\begin{itemize}
\item $\mathbb{V}$ is a set of nodes $s_o^t$ for $o\in \mathbb{O}$ and $t\in \{t_1,\dots,t_k\}$;
\item $\mathbb{E}$ is a set of directed edges of two kinds:
(i) for every 
{collision} $\langle i,j,t \rangle \in \mathbb{C}$, there are two {\em horizontal edges} between nodes $s_i^t$ and $s_j^t$, and
(ii) for every object $o\in \mathbb{O}$ and every consecutive frames $t_i$ and $t_{i+1}$ in $[t_1,\dots,t_k]$, there is a {\em vertical edge} from node $s_o^{t_i}$ to node $s_o^{t_{i+1}}$.
\end{itemize}
Intuitively, each horizontal edge denotes 
{a collision} between the two objects, while each vertical edge denotes the state change over time of the same object.

Given a causal graph $\langle \mathbb{V}, \mathbb{E} \rangle$ constructed for video ${\cal V}$, we formalize causal relationships among nodes in $\mathbb{V}$ as follows. 

\begin{definition}[Ancestor] \label{def:ancestor}
For any two different nodes $s_o^{t}$ and $s_{o'}^{t'}$ in the causal graph,
if there is a path from $s_o^{t}$ to $s_{o'}^{t'}$, we say $s_o^{t}$ is an {\em ancestor} of $s_{o'}^{t'}$. 
\end{definition}

The ancestor relation is used to determine whether the state of one object affects the state of another object and whether the 
{intervention of some} objects affects other objects in subsequent frames.

\begin{definition}[Affected]\label{def:affected}
We say a node $s_o^{t}\in \mathbb{V}$ is {\em affected} by the intervened objects $\mathbb{O}_c$ if (i) $o\in \mathbb{O}_c$ or (ii) there exists a node $s_{o'}^{t'}\in \mathbb{V}$ that is an ancestor of $s_o^{t}$ and $o'\in \mathbb{O}_c$.
\end{definition}

For any object $o\in \mathbb{O}_u$ and any frame $t$, if $s_o^t$ is affected by $\mathbb{O}_c$, 
the simulation for $o$ must start by frame $t$ at the latest. We define the notion of a ``simulation node'' that indicates such a node in the causal graph.

\medskip
\begin{definition}[Sim Node]\label{def:sim-node}
For any node $s_o^{t}\in \mathbb{V}$, it is a {\em simulation node} (or {\em sim node} in short) if (i) $o\in \mathbb{O}_u$, (ii) $s_o^{t}$ is affected by $\mathbb{O}_c$, and (iii) there is no node $s_o^{t'}\in \mathbb{V}$ such that $s_o^{t'}$ is affected by $\mathbb{O}_c$ and $t'<t$.
\end{definition}

\begin{figure}
	\begin{center}
		\centerline{\includegraphics[width=0.6\linewidth,height=2cm]{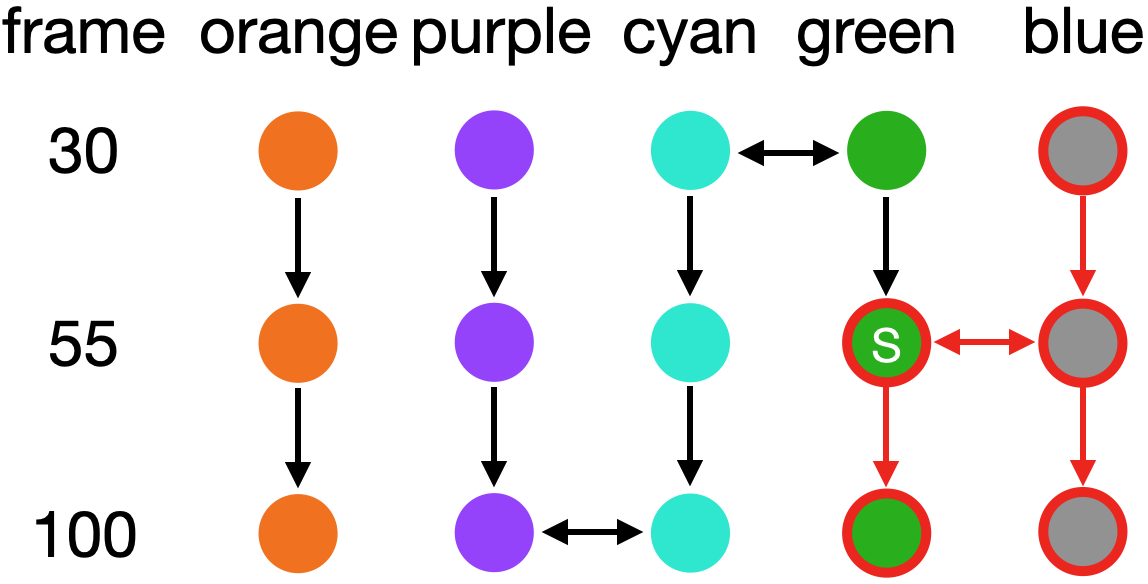}}
		\caption{The causal graph for Example~\ref{ex:causal_graph}. The removed object $blue$ is colored grey. Affected nodes are circled in red. The sim node is denoted by ``s''.}
		\label{fig:graph_clevrer2} 
	\end{center}
\vspace{-2.5em}
\end{figure}

\begin{example}\label{ex:causal_graph}
Figure~\ref{fig:graph_clevrer2} shows an example causal graph for a video, from which ${\cal M}_p$ detects 5 objects $\mathbb{O}=\{orange, purple, cyan, green, blue\}$ and 3 collisions 
$\mathbb{C} = \{\langle cyan,green,30 \rangle, \langle green,blue,55 \rangle, \langle purple,cyan,100 \rangle \}$. 
$\mathbb{V}$ consists of node $s_o^t$ for $o\in \mathbb{O}$ and $t\in \{30, 55, 100\}$.
$\mathbb{E}$ consists of horizontal and vertical edges, as illustrated. 
{Specifically, $\mathbb{E}$ includes horizontal edges between cyan and green object states at frame 30, green and blue object states at 55, and the purple and cyan object states at frame 100. }
The node $s_{blue}^{30}$ is an ancestor of the nodes $s_{blue}^{55}$, $s_{blue}^{100}$, $s_{green}^{55}$, $s_{green}^{100}$, and all are affected by the removed object $\mathbb{O}_c = \{blue\}$. 
Of these five nodes, only $s_{green}^{55}$ satisfies the conditions for being a simulation node. (1) It is not removed, (2) it is affected by some object, and (3) it is the first node $s_{green}^{t}$ to be affected.
\end{example}

\subsection{Enhancing Simulation with Causal Graphs} \label{ssec:crcg}

If we know that node $s_o^t$ is a sim node, we can potentially improve the accuracy of the simulation by replacing the preceding simulated states $\hat{s}_o^i$ with the perception states $s_o^i$ for $i\in\{1,\dots,t-1\}$.
This is the idea behind the Enhanced Simulation module in Figure~\ref{fig:our_model}.
The Enhanced Simulation module, denoted by ${\cal M}_{es}$, takes as input the causal graph $\langle \mathbb{V}, \mathbb{E} \rangle$ constructed in the Causal Graph Construction module, a simulation model ${\cal M}_s$, and the perception states $s_o^t$ of all remaining objects $o\in\mathbb{O}_u$ in all video frames $t\in\{1,\dots,max_v\}$. It then outputs their simulated states and 
{collisions:}
\begin{align*}
&\{[\hat{s}_o^1,\dots,\hat{s}_o^{max_s}]\mid o\in \mathbb{O}_u\}, ~ \hat{\mathbb{C}} \leftarrow \\
&\hspace{10mm}{\cal M}_{es}(\langle \mathbb{V}, \mathbb{E} \rangle, {\cal M}_s, ~\{[s_o^1,\dots,s_o^{max_v}]\mid o\in \mathbb{O}_u\}) 
\end{align*}
where $max_s$ is the maximum number of frames to simulate, and $\hat{\mathbb{C}}$ is the set of 
{collisions} detected in the enhanced simulation, which will be used in the Symbolic Executor module (shown in Figure~\ref{fig:our_model}) to find the answer to the counterfactual question.

\begin{algorithm}[ht!]
\caption{Enhanced Simulation ${\cal M}_{es}$}
\label{alg:enhanced_sim}
\noindent {\bf Input: } 
A causal graph $\langle \mathbb{V}, \mathbb{E} \rangle$ (includes sim nodes), a simulator ${\cal M}_s$, and
(perception) states $[s_o^1,\dots,s_o^{max_v}]$ for $o\in \mathbb{O}_u$ 

\noindent {\bf Output: } States $[\hat{s}_o^1,\dots,\hat{s}_o^{max_s}]$ for $o\in \mathbb{O}_u$, collisions 
$\hat{\mathbb{C}}$

\begin{algorithmic}[1]
\State $\mathbb{O}_p \leftarrow \mathbb{O}_u$;  ~~$\hat{s}_o^1 \leftarrow s_o^1$ for $o\in \mathbb{O}_u$
\ForEach{$t\in \{1,\dots,max_s\}$}
\State $\{\hat{s}_o^{t+1} \mid o\in \mathbb{O}_u\}, ~ \hat{\mathbb{C}}^t \leftarrow {\cal M}_s(\{\hat{s}_o^{t} \mid o\in \mathbb{O}_u\})$ 
\ForEach{$o\in \mathbb{O}_p$}
\If{$t=max_v$ {\bf or} $s_o^t$ is a sim node {\bf or} $\langle o,o',t \rangle\in \hat{\mathbb{C}}^t$ for some $o'\in \mathbb{O}_u\setminus \mathbb{O}_p$}
\State
$\mathbb{O}_p \leftarrow \mathbb{O}_p \setminus \{o\}$
\Else
\State $\hat{s}_o^{t+1} \leftarrow s_o^{t+1}$
\EndIf
\EndFor
\EndFor
\State
\Return $\{[\hat{s}_o^1,\dots,\hat{s}_o^{max_s}]\mid o\in \mathbb{O}_u\}$, ~~ $\hat{\mathbb{C}}=\bigcup\limits_{t=1}^{max_s} \hat{\mathbb{C}}^t$
\end{algorithmic}
\end{algorithm}


Algorithm~\ref{alg:enhanced_sim} describes the detailed procedure in ${\cal M}_{es}$.
In this algorithm, a set of objects $\mathbb{O}_p$ is maintained, where the simulated states of these objects can be replaced with their perception states.
To explain the algorithm's process, let us consider a single frame $t$.
First, Algorithm~\ref{alg:enhanced_sim} calculates the simulated state $\hat{s}_o^{t+1}$ for each object $o\in \mathbb{O}_u$ (line 3).
Next, it checks each object $o\in \mathbb{O}_p$ to see if its perception state is not usable from $t+1$ (line 5).
If this is true, the object is removed from $\mathbb{O}_p$ (line 6).
Otherwise, the algorithm replaces the simulated state $\hat{s}_o^{t+1}$ of the object with its perception state $s_o^{t+1}$ (line 8).


Finally, the symbolic executor of our model (as depicted in Figure~\ref{fig:our_model}) determines the answer to a counterfactual question $Q=\langle \mathbb{O}_c, o_1,o_2 \rangle$. It answers {\em yes} if there exists a {frame} $t$ such that $\langle o_1,o_2,t \rangle\in \hat{\mathbb{C}}$, and {\em no} otherwise.

\begin{figure}[ht!]  
	\begin{center}
		\centerline{\includegraphics[width=\linewidth]{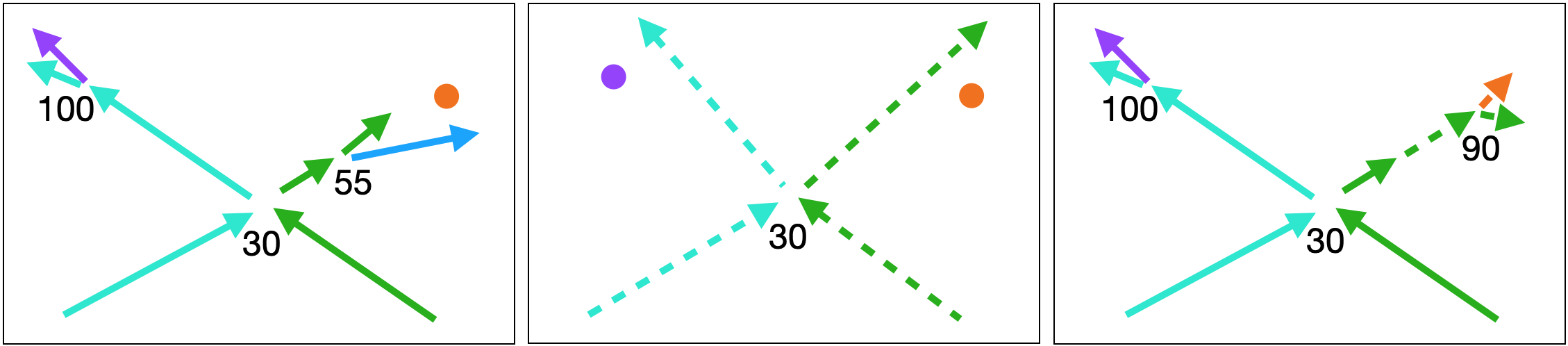}}
		\caption{Object trajectories from the location information. The numbers are the frame numbers for the collisions (left) in the perception states by ${\cal M}_p$; (middle) in the simulated states by ${\cal M}_s$ where simulation starts from the beginning after removing  the blue object; (right) in the enhanced simulation by ${\cal M}_{es}$ using the causal graph. Solid and dashed lines represent trajectories extracted from perception and simulated states, respectively. {The left misses to detect the collision between green and orange; the middle misses to detect two collisions; the right found all three collisions. }}
\label{fig:clevrer:trajectory_example} 
	\end{center}
\vspace{-2em}
\end{figure}

\begin{figure*}
\centering
\includegraphics[width=0.7\textwidth,height=3.6cm]{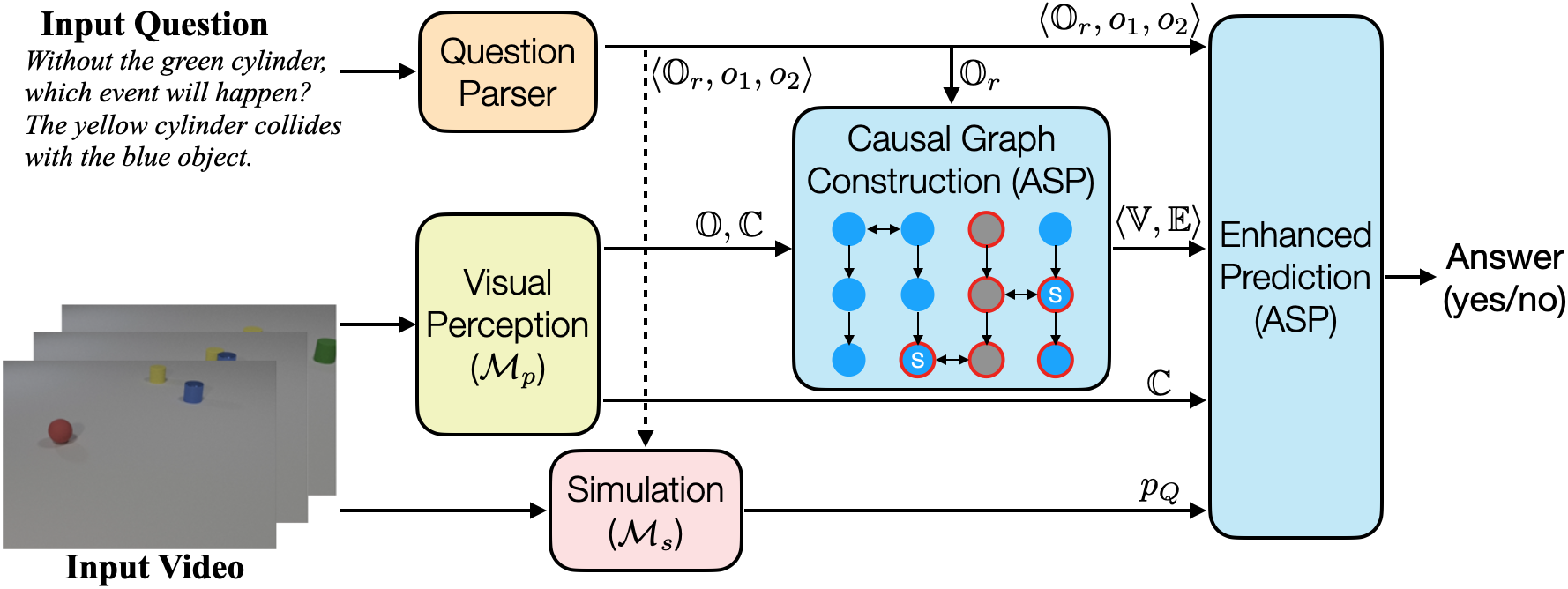}
\caption{Overview of the approximation of $\ours$ when no access to a frame-by-frame simulator is available.} 
\label{fig:asp-ep-pipeline}
\end{figure*}

\noindent{\bf Example~\ref{ex:causal_graph} Continued}\ \  
{\sl 
{When the $blue$ object in Figure~\ref{fig:graph_clevrer2} is removed, ${\cal M}_{es}$ computes the simulated states for the remaining four objects. 
Figure~\ref{fig:clevrer:trajectory_example} visualizes the trajectories of all objects in the original video (left) and the trajectories of the unremoved 
objects in the base (middle) or enhanced (right) simulation. Consider the enhanced simulation (right).
For $o\in \{purple, cyan\}$, the simulated states are the same as the perception states, since there states are not affected. }

Now, suppose that we are asked whether $cyan$ and $purple$ collide. The enhanced simulation detects this collision at frame 100, just as in the original video. However, the basic simulation fails due to the accumulated simulation error. Note that although this collision happens much later than the collision of the removed object, it is still not affected, according to our causal graph in Figure~\ref{fig:graph_clevrer2}.

Finally, let's consider a query about whether $green$ and $orange$ collide. Thanks to the corrected trajectory of $green$ by our enhancement, the enhanced simulation detects this collision at frame 90.}

\subsection{Approximation of $\ours$ When {Frame-by-Frame} Simulator is Not Available} \label{ssec:crcg-approx}

Algorithm~1 assumes that the simulation model ${\cal M}_s$ can perform frame-by-frame simulation. 
It does not apply if ${\cal M}_s$ is a blackbox that could only return the final prediction. 
However, even if ${\cal M}_s$ is a black box that doesn't give intermediate frame results and can only make a final counterfactual predictions  $yes$ or $no$, there is a way to enhance the accuracy by using the information from the causal graph. 
In the following, we design an approximation of $\ours$ with a blackbox simulator, where we identify cases where it is appropriate to reason about the perception result in place of simulation by consulting the causal graph. 

To achieve this, we introduce a new pipeline denoted by $\ours^{approx}$, which is shown in Figure~\ref{fig:asp-ep-pipeline}. Unlike the previous section, here we restrict a counterfactual question $Q=\langle \mathbb{O}_r, o_1, o_2 \rangle$ to be  a special case  ``if we {\em remove} 
some objects $\mathbb{O}_r$ ($\subseteq \mathbb{O}$), will there be {a collision} between objects $o_1$ and $o_2$?'' as in CLEVRER. 

\begin{definition}[Determined]  \label{def:determined}
Consider a counterfactual question $Q=\langle \mathbb{O}_r, o_1, o_2 \rangle$ and a causal graph constructed on objects $\mathbb{O}$ and collisions 
$\mathbb{C}$. 
\begin{itemize}
\item The result of $Q$ is {\em determined to be yes} if there exists some $t$ such that $\langle o_1,o_2,t\rangle \in \mathbb{C}$ and $s_{o_1}^{t}$ and $s_{o_2}^{t}$ are not affected. 
\item  The result of $Q$ is {\em determined to be no} if 
(i) $o_1$ or $o_2$ is in $\mathbb{O}_r$ 
or (ii) $s_{o_1}^{t}$ and $s_{o_2}^{t}$ are not affected and $\langle o_1,o_2,t\rangle \not\in \mathbb{C}$ for any $t$.
\end{itemize}
\end{definition}

Intuitively, Definition~\ref{def:determined} says that the result of a counterfactual question is determined to be yes if the collision happened in the video and the state of the two queried objects at the moment of the collision is not affected by the removed objects. The result is determined to be no if either (i) a queried object in the collision is removed, or (ii) the collision did not happen in the video and all the states of the queried objects are not affected by the removed objects.

\begin{figure}[h!]
\begin{center}
\centerline{\includegraphics[width=\linewidth,height=1.53cm]{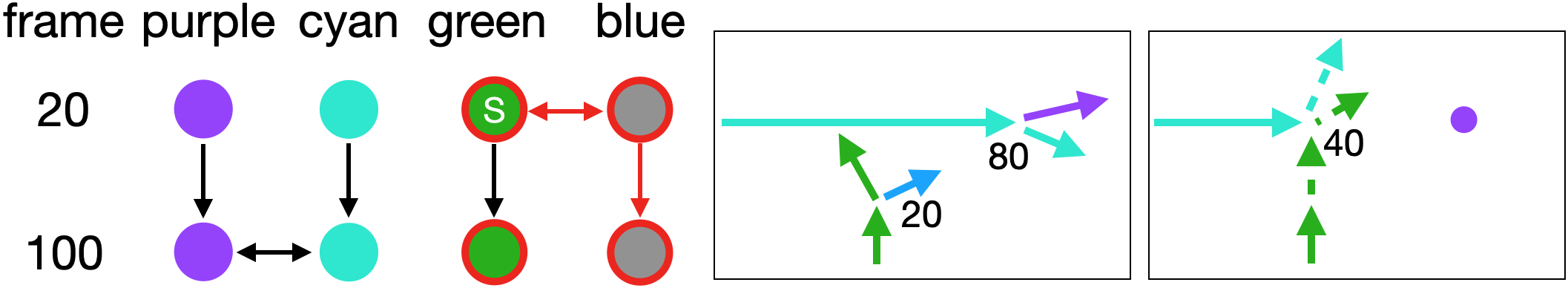}}
\caption{(left) The causal graph constructed for a video with 4 objects and 2 collisions. On its right are object trajectories from the location information in the perception states by ${\cal M}_p$ (middle) or in the simulated states by ${\cal M}_{es}$ (right).} 
\label{fig:ex_approx} 
\end{center}
\vspace{-2em}
\end{figure}

\smallskip
\noindent{\bf Remark~}
Note that, unlike Algorithm~\ref{alg:enhanced_sim}, it is inevitable that the above-determined result cannot capture the influence from other simulated objects and thus is not guaranteed to be always correct even with a perfect perception model ${\cal M}_p$. 
Consider the example in Figure~\ref{fig:ex_approx} where the counterfactual question is $Q=\langle \{blue \}, purple, cyan\rangle$. While its result is determined to be yes according to the above definition, the collision between $purple$ and $cyan$ shouldn't happen because $green$ would collide with $cyan$ if $blue$ were removed, which changes the trajectory of $cyan$ so that $cyan$ wouldn't hit $purple$. Such examples are possible, yet don't occur often in practice.\footnote{See Table~\ref{table:clevrer:cf:ablation} for illustration.}

\smallskip

In the end, given a baseline counterfactual prediction $p_Q\in\{yes, no\}$, the Enhanced Prediction module in Figure~\ref{fig:asp-ep-pipeline} gives the final answer on a counterfactual question 
$Q=\langle \mathbb{O}_c, o_1, o_2 \rangle$: 
if the result is determined to be yes or no, the final answer is the same. Otherwise (i.e., if the result is not determined), the final answer is the same as $p_Q$.
%
In other words, the determined fact obtained from the perception states overrides the baseline's prediction $p_Q$. 
This can be understood as an approximation of Algorithm~\ref{alg:enhanced_sim} where $max_s=max_v$ and the checking of ``$\langle o,o',t \rangle\in \mathbb{C}^t$ for some $o'\in \mathbb{O}_u\setminus \mathbb{O}_p$'' in 
{line} 5 is removed.

\section{Realization of $\ours$ in ASP}\label{sec:asp-impl}

This section models the causal graph using answer set programming and uses an answer set solver to derive the necessary information. 


Given a video ${\cal V}$, a perception model ${\cal M}_p$ returns the in-video collisions
$\mathbb{C}$ and the perception states of all objects $\mathbb{O}$. 
These outputs {and a counterfactual question $Q=\langle \mathbb{O}_c, o_1, o_2 \rangle$} are represented in the ASP facts as follows. 
{We represent each collision $\langle i,j,t \rangle$ in $\mathbb{C}$ with an {\em atom} {\tt collision(i,j,t)}, and represent each object $o\in\mathbb{O}$, $o\in\mathbb{O}_c$, and $o\in\mathbb{O}_r$ with atoms {\tt object(o)}, {\tt intervened(o)}, and {\tt removed(o)} respectively.}
The atoms {\tt ancestor(o,t,o',t')}, {\tt affected(o,t)}, and {\tt sim(o,t)} represent ``$s_o^{t}$ is an ancestor of $s_{o'}^{t'}$,'' ``$s_o^t$ is affected (by the 
{intervened} objects),'' and ``$s_o^t$ is a sim node,'' respectively.
We designed an ASP program $\Pi$ to represent the causal graph and to deduce a single answer set $\mathbb{A}$ such that 
(i) {\tt ancestor(o,t,o',t')} $\in \mathbb{A}$ iff $s_o^{t}$ is an ancestor of $s_{o'}^{t'}$, 
(ii) {\tt affected(o,t)} $\in \mathbb{A}$ iff $s_o^{t}$ is affected, 
and (iii) {\tt sim(o,t)} $\in \mathbb{A}$ iff $s_o^{t}$ is a sim node.

The definitions introduced in Section~\ref{sec:math_model} can be represented in ASP in a straightforward way.

\medskip
\noindent{\bf Ancestor~~}
The ancestor relation is introduced either by the same object with different frames (i.e., vertical edges in the causal graph)
\begin{lstlisting}
ancestor(O,T1,O,T2) :- object(O),
               collision(_,_,T1), collision(_,_,T2), T1<T2.
\end{lstlisting}
or by a collision 
(i.e., horizontal edges in the causal graph)
\begin{lstlisting}
ancestor(O1,T,O2,T) :- collision(O1,O2,T).
collision(O1,O2,T) :- collision(O2,O1,T).
\end{lstlisting}
or by the transitive closure of the ancestor relation itself.
\begin{lstlisting}
ancestor(O1,T1,O2,T2) :- ancestor(O1,T1,O3,T3),
                   ancestor(O3,T3,O2,T2), (O1,T1)!=(O2,T2).
\end{lstlisting}
In ASP, a variable starts with an uppercase letter, a constant starts with a lowercase letter, and an underscore represents an anonymous variable. The term {\tt (O1,T1)!=(O2,T2)} in the last rule guarantees that nodes $s_{O1}^{T1}$ and $s_{O2}^{T2}$ are not the same, thus no edge will be introduced to form a self-loop.

\medskip
\noindent{\bf Affected~~}
A node $s_o^t$ is affected if $o$ is intervened 
\begin{lstlisting}
affected(O,T) :- intervened(O), collision(_,_,T).
\end{lstlisting}
or there is an 
{intervened object $o'$} such that $s_{o'}^{t'}$ is an ancestor of $s_o^t$ for some $t'$.
\begin{lstlisting}
affected(O,T) :- intervened(O'), ancestor(O',T',O,T).
\end{lstlisting}

\medskip
\noindent{\bf Sim Node~~}
Node $s_o^t$ is a sim node if 
object $o$ is not removed, node $s_o^t$ is affected,
and no node $s_o^{t'}$ in earlier frame $t'$ is affected.
\begin{lstlisting}
sim(O,T) :- not removed(O), affected(O,T), 
            T<=T': affected(O,T').
\end{lstlisting}
In the above rule, {\tt T<=T': affected(O,T')} is a {\em conditional literal} saying that ``for all cases when node $s_{O}^{T'}$ is affected, $T'$ must be greater or equal to $T$.''

\smallskip
\noindent
{\bf Implementation of $\ours$ (Sec~\ref{ssec:crcg}) using ASP}\ \ \ Assuming the availability of the frame-by-frame simulator ${\cal M}_s$, 
we implement Algorithm~\ref{alg:enhanced_sim} in ``Enhanced Simulation Model (${\cal M}_{es}$)'' (Figure~\ref{fig:our_model}) to find all collisions $\hat{\mathbb{C}}$ using the perception states detected from ${\cal M}_p$ and the sim node information {\tt sim(O,T)} (encoding $s_o^t$ is a sim node) computed by ASP. 
For counterfactual question $Q=\langle \mathbb{O}_c, o_1, o_2 \rangle$,  the answer is {\tt yes} (i.e., the event in $Q$ happens) if a collision between $o_1$ and $o_2$ belongs to $\hat{\mathbb{C}}$. Otherwise, the answer is {\tt no}. 

\section{Experiments with CLEVRER}\label{sec:clevrer-experiment} 

\subsection{Applying CRCG to CLEVRER} 

We utilized our approach, $\ours$, in conjunction with both the perception model and the simulation model within VRDP \cite{ding2021dynamic}. Specifically, we refer to the combination of $\ours$ with the VRDP model as $\ours_{VRDP}$. Additionally, we applied $\ours^{approx}$ to enhance the predictions of NS-DR directly, and we denote this combination as $\ours^{approx}_{NSDR}$.

\begin{table}[ht!]
	\small
	\caption{Performance comparison with the state-of-the-art methods on counterfactual questions in CLEVRER test set.}
	\centering
		\begin{tabular}{l|cc} 
		\toprule
		Model & Opt.Accuracy (\%) & Ques. Accuracy (\%)\\
		\midrule
		NS-DR \cite{yi19clevrer} & 74.1 & 42.2 \\
		DCL \cite{chen21grounding} & 80.4 & 46.5 \\
		Aloe \cite{ding21attention} & 91.4 & 75.6 \\
		ODDN-Aloe \cite{tang22object} & 93.0 & 80.1 \\
		VRDP \cite{ding2021dynamic} & 94.8 & 84.3 \\
		\midrule
        $\ours_{NSDR}^{approx}$ & 90.7 & 78.3 \\
		$\ours_{VRDP}$ & {\bf 96.1} & {\bf 87.8} \\
		\bottomrule
		\end{tabular}
	\label{table:clevrer:cf}
\end{table}

Table~\ref{table:clevrer:cf} demonstrates the significant improvement in option accuracy and question accuracy resulting from our enhancement of NS-DR predictions. Furthermore, $\ours_{VRDP}$ outperforms the VRDP baseline and establishes a new state-of-the-art level of performance for counterfactual questions 
in the CLEVRER dataset. We measure the average runtime of the ASP programs computing the causal graphs to be 6 milliseconds.

\begin{table}[ht!]
	\small 
	\caption{Ablation study on CLEVRER validation set of how the source (i.e., perception or simulation) of predicted collision for an option affects the option accuracy. The number of options and accuracy are reported for each source.}
		\centering	
		\begin{tabular}{l|cc}	
		\toprule	
		\multirow{2}{*}{Model} & determined & not determined\\	
		& \#opt. / acc(\%) & \#opt. / acc(\%) \\	
		\midrule	
		NS-DR                   & 20558 / 75.31 & 12493 / 71.78 \\	
		$\ours_{NSDR}^{approx}$ & 20558 / 95.08 & 12493 / 71.78 \\	
		\bottomrule	
		\end{tabular}	
	\label{table:clevrer:cf:ablation}	
\end{table}

In the CLEVRER dataset, every question-option pair can be represented as a tuple $\langle \mathbb{O}_c, o_1, o_2 \rangle$. To evaluate the effectiveness of $\ours^{approx}$ in improving baselines, we have divided these tuples into two sets based on whether the answer is {\tt determined} or not. The improvements on each set are shown in Table~\ref{table:clevrer:cf:ablation}. We have observed that $\ours^{approx}_{NSDR}$ performs much better in the first column. The reason is that collisions detected directly from the video using ${\cal M}_p$ in NS-DR are generally more accurate than the collisions predicted using ${\cal M}_s$ simulation from frame 1 onwards. Applying $\ours$ significantly enhances the option accuracy from 75.31\% to 95.08\% for those options where in-video collisions can be used. However, the accuracy remains the same for the remaining options, as both methods use the same baseline prediction $p_Q$.

\subsection{Enhancements for Other Question Types of CLEVRER}

This section describes how we can further enhance neuro-symbolic models for other question types in CLEVRER, leveraging their modular architecture. Table~\ref{tab:overall} presents a comparison of different models' performance on the CLEVRER test set, including NS-DR and VRDP models improved by our approach.

To achieve state-of-the-art performance on all question categories, we designed two additional modules to improve the accuracy of perception and simulated states, respectively. 

\smallskip
\noindent{\bf Improved Object Detection (IOD)~~}
IOD is a post-processor for the perception model ${\cal M}_p$ that reduces its output noise and errors through two functions: {\em trajectory smoothing} and 
{\em topmost as center}. 
{\em Trajectory smoothing} draws a virtual line to connect the trajectories and interpolates missing frames. This helps reduce noise and errors in the output. 
In addition,  since there are missing trajectories of objects at some frames due to occlusion and errors, 
{\em topmost as center} uses the topmost point (instead of center) of an object to trace its trajectory. This is because the topmost position is less likely to be occluded than the center point.

\smallskip
\noindent{\bf Simple Physics Simulator (SPS)~~}
We used the smoothed trajectories generated by IOD to create a basic physics simulator, which serves as ${\cal M}_s$ for predicting and exploring counterfactual scenarios. Our simulator represents the motion of objects in a two-dimensional space, where each object is treated as a point and follows simple kinematic equations to determine its movement.

\begin{table}
	\caption{
Performance of models among all question categories on CLEVRER test set. Also, refer to the leaderboard \url{https://eval.ai/web/challenges/challenge-page/667/leaderboard/1813}.}
	
\footnotesize 
\setlength{\tabcolsep}{2pt}
\begin{tabular}{|l|l|l|l|l|l|l|l|l|}
\hline
\multirow{2}{*}{{ Model}}  & \multicolumn{2}{l|}{\multirow{2}{*}{Desc.}} & \multicolumn{2}{l|}{Explanatory} & \multicolumn{2}{l|}{Predictive} & \multicolumn{2}{l|}{Counterfact.} \\ \cline{4-9} 
& \multicolumn{2}{l|}{}           &  opt.     &  ques.  &  opt.  &   ques.   &  opt.   &  ques.   \\ \hline
NS-DR \cite{mao19neuro}   & \multicolumn{2}{l|}{88.1}              &   87.6        &     79.6      &     82.9      &    68.7       &     74.1      &    42.2       \\ \hline
DCL \cite{chen21grounding}       & \multicolumn{2}{l|}{90.7}       &   89.6        &     82.8      &     90.5      &    82.0       &     80.4      &    46.5       \\ \hline
Aloe \cite{ding21attention}    & \multicolumn{2}{l|}{94.0}       &   98.47       &     96.0      &     {93.5}    &    87.5       &     {91.42}   &    75.61       \\ \hline
VRDP \cite{ding2021dynamic}    & \multicolumn{2}{l|}{93.4}         &   96.3        &     91.9      &     95.7      &    91.4       &     {94.8}      &    84.3       \\ \hline
ODDN-Aloe \cite{tang22object}    & \multicolumn{2}{l|}{95.8}         &     98.9      &     97.0      &    95.7       &     {91.8}      &    93.0   & 80.1     \\ \hline
$\ours^{approx}_{NSDR}$+IOD+SPS    & \multicolumn{2}{l|}{{95.55}}                 &\textbf{99.94} &\textbf{99.81} &   88.12       &   {76.64}      &   90.73       &   {78.31}   \\ \hline
$\ours_{VRDP}$+IOD+SPS    & \multicolumn{2}{l|}{\textbf{96.46}}           &   {{99.32}}   &{{98.80}}      & \textbf{96.11}& \textbf{92.28} &   \textbf{96.61}       &   \textbf{90.72}   \\
\hline 
\end{tabular}
\label{tab:overall}
\end{table}

\section{Experiments with CRAFT}\label{sec:craft-experiment}

The CRAFT dataset~\cite{ates20craft} contains 10,000 videos that depict causal relationships between falling and sliding objects. It also includes 57,000 questions. CRAFT differs from CLEVRER by introducing additional events and presenting many different environments that feature various configurations of immovable objects such as ramps and baskets. The questions in the dataset are divided into three categories, but we only consider counterfactual questions.

\begin{figure}[ht]  
\begin{center}
\centerline{\includegraphics[width=1.0\linewidth]{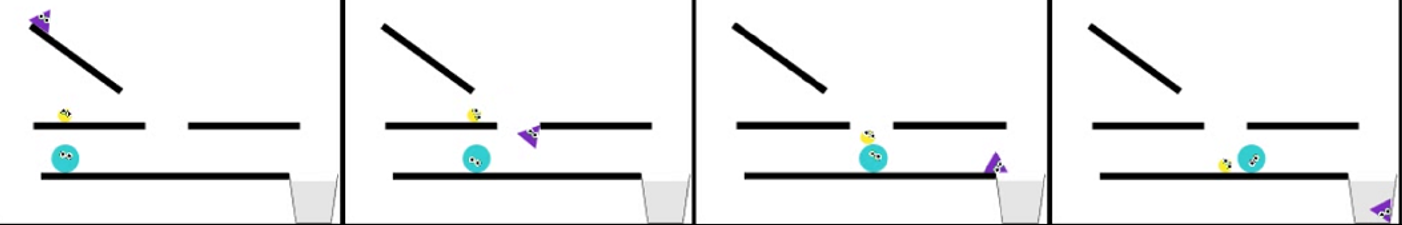}}
\caption{Screenshots from sample CRAFT video \#1,179.}
\label{fig:sample_craft_video} 
\end{center}
\vspace{-5mm}
\end{figure}

As there are no baselines for CRAFT that provide the necessary intermediate information for constructing causal graphs -- such as object static features (e.g. color, shape, size) and in-video collision information -- we explore an alternative approach. Specifically, we take advantage of GPT-x, a large language model, as a proxy for a simulator (denoted as ${\cal M}_s$ in Figure~\ref{fig:asp-ep-pipeline}), by utilizing the textual descriptions of the videos available within the CRAFT dataset.
For example, a description-query pair for video \#1,179 (shown in Figure~\ref{fig:sample_craft_video}) from the test set is:
\begin{lstlisting}
Start. Large cyan circle collides with small yellow circle. Small purple triangle enters basket. Large cyan circle collides with small yellow circle. Small purple triangle collides with basket. End.

Will the tiny purple triangle end up in the basket if the large cyan circle is removed?
\end{lstlisting}

We compare three settings: vanilla GPT-x, $\ours^{approx}$ using GPT-x as a proxy simulator, and another GPT-x case where we give a modified prompt by leveraging the ``determined" information obtained as part of $\ours^{approx}$. \footnote{{The GPT-3.5 model used in our experiments is ``gpt-3.5-turbo-0613'', the GPT-4 model used is ``gpt-4-0613'' and the temperature is set to 0 for all experiments.}}

\smallskip
\noindent{\bf GPT-x~~}
The baseline GPT-x model receives two inputs: natural language descriptions of the scenes in the CRAFT dataset, and a prompt generated from video descriptions and a question. It then outputs an answer, such as ``Yes'' or ``No''. 


\smallskip
\noindent{\bf $\ours_{GPTx}^{approx}$ ~~} 
%
%
%
Since we input natural language descriptions of the scenes, we extend the question parser to extract symbolic information, such as removed objects $\mathbb{O}_c$ and in-video collisions $\mathbb{C}$ from both the question and the video description. Subsequently, the ASP implementation of $\ours^{approx}$ converts this information into a causal graph and determines whether the result of the question is {\tt determined}. If it is, in-video collisions $\mathbb{C}$ are used to answer the question. Otherwise, we resort to the baseline GPT-x prediction as in the first setting. 

\smallskip
\noindent{\bf GPT-x with $\ours$ guided prompt~~} 
Similar to the second setting, the process first checks if the result is {\tt determined}. 
If not, we default to the baseline GPT-x prediction as in the first setting.
Conversely, if the result is {\tt determined}, we rephrase the given counterfactual question (e.g., "Will the blue circle fall to the ground if the large blue triangle is removed?") as a perception question (e.g., ``Did the small blue circle fall to the floor?''), and then use GPT-x to answer it. 

\begin{table}[h!]
	\small
	\caption{\small Accuracy on counterfactual questions in CRAFT test dataset} 
 \vspace{-1mm}
	\centering
		\begin{tabular}{@{}l|cccccc@{}}
			\toprule
			\multirow{2}{*}{Model}  & \multirow{1}{*}{Easy Split } & \multirow{1}{*}{Hard Split }\\ 
			\multirow{1}{*}{}  & \multirow{1}{*}{Ques.(\%)} & \multirow{1}{*}{Ques.(\%)}\\  \midrule
			LSTM-D \cite{ates20craft} & 55.89    &   56.00      \\  
			BERT-D \cite{ates20craft} & 80.05    &   79.34      \\
			\hline
   			GPT-3.5                       & 52.29    &   52.53      \\
			$\ours_{GPT3.5}^{approx}$        & 65.32    &   68.48      \\
			GPT-3.5 with $\ours$ guided prompt            & 63.26    &   65.89      \\
                \bottomrule
   			GPT-4                       & 77.93    &   81.20      \\
			$\ours_{GPT4}^{approx}$        & 79.68    &   83.64      \\
			GPT-4 with $\ours$ guided prompt            & 78.07    &   81.22      \\
			\bottomrule
		\end{tabular}
	\label{tab:CRAFT_3_2_all} 
 \vspace{-2mm}
\end{table}

\medskip
Table~\ref{tab:CRAFT_3_2_all} compares the performance of the aforementioned cases with the LSTM-D and BERT-D models from~\cite{ates20craft}. The comparison is based on both the "easy" and "hard" splits of counterfactual questions in the CRAFT test set.
While LSTM-D and BERT-D also take questions and textual descriptions of scenes as input, they need to be trained on approximately 34K training instances. In contrast, all GPT-x-based methods are in a few-shot setting.

To better understand the effectiveness of $\ours$, we compared the test accuracy on 1,128 counterfactual questions (out of 3,489) whose results were {\tt determined}. The results are presented in Table~\ref{tab:CRAFT_3_2_branch2}. {The $\ours_{GPTx}^{approx}$ models achieved an impressive 97.96\%/99\% accuracy on the easy/hard split, which is much higher than their respective GPT-x baselines.} Notably, using {\tt determined} to guide the GPT-x prompt also improved the baseline accuracy by about {13\% for GPT-3.5}, demonstrating that prompt engineering can greatly benefit from the causal graph. Note that $\ours_{GPT3.5}^{approx}$ and $\ours_{GPT4}^{approx}$ have the same performance, this is due to the fact that for the determined cases, the solution is found without using the simulator, thus in both cases no simulator is used and the ASP programs are identical.

\begin{table}
	\small
	\caption{Accuracy on counterfactual questions in CRAFT test dataset whose results are {\tt determined}.}
	\centering
		\begin{tabular}{@{}l|cccccc@{}}
			\toprule
			\multirow{2}{*}{Model}  & \multirow{1}{*}{Easy Split } & \multirow{1}{*}{Hard Split }\\ 
			\multirow{1}{*}{}  & \multirow{1}{*}{Ques.(\%)} & \multirow{1}{*}{Ques.(\%)}\\  \midrule
   			GPT-3.5                       & 57.62    &   53.99      \\
			$\ours_{GPT3.5}^{approx}$        & 97.96    &   99.00    \\ 
			GPT-3.5 with $\ours$ guided prompt            & 91.58    &   91.69      \\  \bottomrule
   			GPT-4                       & 92.55    &   92.11      \\
			$\ours_{GPT4}^{approx}$        & 97.96    &   99.00    \\ 
			GPT-4 with $\ours$ guided prompt            & 93.00    &   92.19      \\  \bottomrule
		\end{tabular}
	\label{tab:CRAFT_3_2_branch2} 
 \vspace{-2mm}
\end{table}

There are other similar benchmarks like CLEVRER and CRAFT that deals with counterfactual reasoning under physics constraints, such as ComPhy \cite{chen22comphy} and Cophy \cite{baradel20cophy}. We believe that our approach is applicable to these datasets as well, but the required frame-by-frame simulator is non-trivial to build and we couldn't find a publicly available one.

\vspace{-1mm}
\section{Conclusion}  \label{sec:conclusion}

In this paper, we present a method for strengthening neuro-symbolic models by 
identifying the relation between actual and counterfactual states via explicit causal reasoning in answer set programming. 
Our method improves upon the baseline as long as perception is more accurate than simulation (which is typically the case), while using the same perception/simulation modules.
As the experimental analysis shows, a further improvement is possible with a more accurate 
{simulation module}; this is again a benefit of the modular architecture.
Moreover, the computation of our approach is interpretable, which led us to improve some modules by augmenting them with additional computation.

\medskip\noindent
{\bf Acknowledgements }
We are grateful to the anonymous referees for their useful comments. This work was partially supported by the National Science Foundation under Grant IIS-2006747.

{\small

}

\clearpage

\appendix

{\Large Appendix to ``Think before You Simulate: Symbolic Reasoning to Orchestrate Neural Computation for Counterfactual Question Answering"}
\bigskip


Section~\ref{appendix:sec:datasets} describes more details about the datasets. 
Section~\ref{appendix:sec:implement_crcg_approx} details implementation of $\ours^{approx}$ in ASP.
Section~\ref{appendix:sec:gpt3} gives the examples of the three methods for the CRAFT experiment (Section~\ref{sec:craft-experiment}).
Section~\ref{appendix:sec:other_enhancements} describes the details of how we achieve the SOTA performance on all four types of the CLEVRER questions accompanied by ablation studies. 
Section~\ref{appendix:sec:asp} presents the full ASP programs we wrote. 

All experiments were done on Ubuntu 18.04.2 LTS with two 10-core CPU Intel(R) Xeon(R) CPU E5-2640 v4 @ 2.40GHz and four GP104 [GeForce GTX 1080]. We use Clingo version 5.3.0. ASP solving time is instant. 

\section{Datasets}\label{appendix:sec:datasets}

\subsection{CLEVRER}
The CLEVRER dataset \cite{yi19clevrer} consists of 20K synthetic videos of colliding objects and more than 300K questions about objects in motion and their interactions. The videos are five seconds with a total of 127 frames and contain various objects moving and colliding with each other.
The questions about the videos are divided into four categories.
{\em Descriptive questions} are about intrinsic attributes of objects, such as color, shape, and material, and the events related to the objects, such as collision and entering/exiting the scenes. 
{\em Explanatory questions} are about if an object or a collision event is a cause for another collision event or an object exiting the scene.
	%
{\em Predictive questions} are about whether a collision will happen  after the video ends. 
	%
{\em Counterfactual questions} are about collisions that would or would not happen if some object in the video were removed. 
Except for descriptive questions, which require short answers, all the other questions are multiple-choice questions requiring one to select all true choices. 
	



\subsection{CRAFT}

The CRAFT dataset~\cite{ates20craft} consists of 10K videos involving causal relations between falling and sliding objects, along with 57K questions. Compared to CLEVRER, CRAFT introduces additional events and has many different environments, including various configurations of immovable objects such as ramps and the basket. 
The three main categories of questions are descriptive, causal, and counterfactual.
We focus only on counterfactual questions, which have 6 types. These question types
are generally more complex than CLEVRER counterfactual questions, sometimes requiring multiple counterfactual simulations (e.g., ``Will the large green square enter the basket if any of the other objects are removed?''), or counting (e.g., ``How many objects fall to the ground if the small blue box is removed?'').

The counterfactual questions in the CRAFT dataset are comprised of 6 types. Examples of each are as follows:

\begin{lstlisting}
Type 1: Removing one object, does enter event happen?
If the small brown triangle is removed, will the big green circle fall into the bucket?
Will the large cyan circle end up in the basket if the tiny cyan triangle is removed?
Will the tiny purple circle fall into the container if the tiny purple triangle is removed?

Type 2: Removing one object, does ground collision event happen?
will the small purple triangle hit the floor if the large gray circle is removed?
will the large gray circle hit the ground if the tiny purple triangle is removed?
will the small purple circle fall to the ground if the big gray circle is removed?

Type 3: Removing one object, how many enter events happen?
How many objects go into the bucket if the tiny gray circle is removed?
If the tiny yellow triangle is removed, how many objects get into the basket?
How many objects get into the basket if the large green triangle is removed?

Type 4: Removing one object, how many ground collision events?
If the small gray circle is removed, how many objects hit the ground?
How many objects fall to the ground if the big red triangle is removed?
If the large brown triangle is removed, how many objects fall to the ground?

Type 5: Removing any object, does enter event happen?
Will the tiny purple triangle fall into the bucket if any of the other objects are removed?
Will the small gray circle fall into the bucket if any of the other objects are removed?
If any of the other objects are removed, will the large yellow triangle get into the basket?

Type 6: Removing any object, does ground collision happen?
If any one of the other objects are removed, will the tiny purple circle fall to the floor?
Will the large blue triangle hit the floor if any of the other objects are removed?
If any one of the other objects are removed, will the big gray cube hit the ground?

\end{lstlisting}

\section{Implementation of $\ours^{approx}$ (Sec~\ref{ssec:crcg-approx}) using ASP}\label{appendix:sec:implement_crcg_approx} 

If the frame-by-frame simulator is not accessible, we use the approximation of Algorithm~1 to answer. For this, we need a few more ASP rules. 
For a counterfactual question $Q=\langle \mathbb{O}_c, o_1, o_2 \rangle$, and its prediction $p_Q\in\{yes,no\}$ from any baseline, 
let's represent the prediction $p_Q$ as {\tt predict(Q, yes)} or {\tt predict(Q, no)} according to its value and represent each collision $\langle i,j,t \rangle$ in $\mathbb{C}$ with {\tt event(i,collide,j,t)}. 

\medskip
\noindent{\bf Determined~~}
The result of a counterfactual question $Q=\langle \mathbb{O}_c, o_1, o_2 \rangle$ is determined to be {\tt yes} if the collision happened in the video at a time when the states of $o_1$ and $o_2$ are not affected by the removed object.  	
\begin{lstlisting}
determined(Q, yes) :- query(Q, qobj(I1), Event, qobj(I2)), 
        same(qobj(I1), O1), same(qobj(I2),O2),
        event(O1, Event, O2, F),
        not affected(O1,F), not affected(O2,F).
\end{lstlisting}
Here,
{\tt query(Q,qobj(1),Event,qobj(2))} represents
``query {\tt Q} is about the {\tt Event} happening between objects $o_1$ and $o_2$,'' and 
{\tt same(qobj(1),O)} represents 
``$o_1$ is the same as the {\tt O}-th object in the video.''

The result is determined to be {\tt no} if $o_1$ or $o_2$ is removed,
\begin{lstlisting}
determined(Q, no) :- query(Q, qobj(I1), Event, qobj(I2)), 
        same(qobj(I1), O1), same(qobj(I2),O2),
        removed(O1): not removed(O2).
\end{lstlisting}
or if the collision didn't happen in the video and the states of $o_1$ and $o_2$ are not affected by the removed objects.
\begin{lstlisting}
determined(Q, no) :- query(Q, qobj(I1), Event, qobj(I2)), 
        same(qobj(I1), O1), same(qobj(I2),O2),
        not event(O1, Event, O2, _),
        not affected(O1,_), not affected(O2,_).
\end{lstlisting}

Then, given a prediction $p_Q$, the answer to $Q$ is {\tt Res} if (i) the result of $Q$ is determined to be {\tt Res}, or (ii) the value of $p_Q$ is {\tt Res}, and the result of $Q$ is not determined.
\begin{lstlisting}
answer(Q, Res) :- determined(Q, Res).
answer(Q, Res) :- predict(Q, Res), not determined(Q, _).
\end{lstlisting} 

The answer to $Q=\langle \mathbb{O}_c, o_1, o_2 \rangle$ is directly obtained from the value of {\tt Res} in the {\tt answer} atoms that are derived.

\section{Example Flow of CRAFT Experiment}\label{appendix:sec:gpt3}


The GPT-x model used in our experiments is ``text-davinci-002'' and the temperature is set to 0 for the reproductivity of all experiments.
A description-query pair for example video \#1,179 (shown in Figure~\ref{fig:sample_craft_video}) from the CRAFT test set is:

\begin{lstlisting}
Start. Large cyan circle collides with small yellow circle. Small purple triangle enters basket. Large cyan circle collides with small yellow circle. Small purple triangle collides with basket. End.

Will the tiny purple triangle end up in the basket if the large cyan circle is removed?
\end{lstlisting}

\subsection{GPT-x Baseline} 

In the GPT-x baseline, a counterfactual question is asked directly in a prompt, consisting of an instruction, a video description (from the CRAFT dataset), and the counterfactual question itself. 
An example prompt for the example in Figure~\ref{fig:sample_craft_video} is shown below.

\begin{lstlisting}
Instructions: A description of a scene of moving objects and their physical dynamics is presented. A question is then asked about hypothetical changes in the scene and their outcomes.

Description: Start. Large cyan circle collides with small yellow circle. Small purple triangle enters basket. Large cyan circle collides with small yellow circle. Small purple triangle collides with basket. End.

Will the tiny purple triangle end up in the basket if the large cyan circle is removed? (yes or no)

\end{lstlisting}
For this example, the GPT-x responds ``No'', which is incorrect.

\subsection{$\ours^{approx}_{GPTx}$: Enhance GPT-x Baseline with $\ours^{approx}$}

We describe the whole process with the above example. From the description-query pair
\begin{lstlisting}
Start. Large cyan circle collides with small yellow circle. Small purple triangle enters basket. Large cyan circle collides with small yellow circle. Small purple triangle collides with basket. End.

Will the tiny purple triangle end up in the basket if the large cyan circle is removed?
\end{lstlisting}
we first use a python script to extract the following atomic facts into an ASP program `{\tt input.lp}'.

\begin{lstlisting}
size(0,large).color(0,cyan).shape(0,circle).
size(1,small).color(1,purple).shape(1,triangle).
size(2,small).color(2,yellow).shape(2,circle).
size(95,large).color(95,black).shape(95,ground).
size(97,large).color(97,black).shape(97,basket).
collision(0,2,0).
collision(0,2,2).
collision(1,97,3).
enter(1,97,1).

counterfact(remove,qobj(0)).
feature(qobj(0),large).
feature(qobj(0),cyan).
feature(qobj(0),circle).

option(1, qobj(1), enter, qobj(2)).
feature(qobj(1),small).feature(qobj(1),purple).feature(qobj(1),triangle).
feature(qobj(2),large).feature(qobj(2),black).feature(qobj(2),basket).

\end{lstlisting}

The background knowledge about moving dynamics is encoded in  `{\tt causal.lp}' which is the one presented in Section~\ref{sec:asp-impl}. The full ASP program is given in Appendix~\ref{appendix:sec:asp}.
The answer set of {\tt input.lp} + {\tt causal.lp} is:
\begin{lstlisting}
sim(2,0), determined(1, yes), answer(1,yes), ...
\end{lstlisting}

The answer set contains the fact \texttt{answer(1,yes)}, meaning that the answer to the query ``Will the tiny purple triangle end up in the basket if the large cyan circle is removed?'' is yes, which is the correct answer.

\subsection{GPT-x with $\ours$ guided prompt}
For GPT-x with $\ours$ guided prompt, when $\ours$ derives {\tt determined(Q,R)} for a counterfactual question {\tt Q}, the prompt for GPT-x is replaced with the following perception question.
\begin{lstlisting}
Instructions: A description of a scene of moving objects and their physical dynamics is presented. A question is then asked about the scene description.

Description: Start. Large cyan circle collides with small yellow circle. Small purple triangle enters basket. Large cyan circle collides with small yellow circle. Small purple triangle collides with basket. End.

According to the scene description, did the tiny purple triangle end up in the basket? (yes or no)

\end{lstlisting}
The GPT-x responds  ``Yes,'' which is correct.

\section{Other Enhancements to Neuro-Symbolic Models for CLEVRER Tasks}\label{appendix:sec:other_enhancements}

\subsection{Improved Object Detection (IOD)}\label{ssec:iod}

For the descriptive and explanatory questions in the CLEVRER dataset, it is important to recognize the events in the video correctly. 
To help improve the accuracy of the detected results by the perception model ${\cal M}_p$, we implement a simple method called IOD (Improved Object Detection).

IOD is a post-processor for the perception model ${\cal M}_p$ to reduce its output noise and errors through two functions: {\em trajectory smoothing} and {\em topmost as center}. 
The input to IOD is the object trajectories output from the perception model ${\cal M}_p$.
Since there are missing trajectories of objects at some frames due to occlusion and errors, {\em trajectory smoothing} draws a virtual line to connect the trajectories and uses interpolation for the frames an object is missing. {\em Trajectory smoothing} is applied to both NS-DR and VRDP, yielding better accuracy on all question types.

In the case of NS-DR, the Mask-RCNN outputs the mask of each object, and PropNet uses the {\em center} of the mask as the object's position.
However, this method has a defect because occlusion could make the center of the mask move abruptly, leading to wrong answers for questions like how many objects are moving. 
To address this issue, the IOD module for NS-DR also applies {\em topmost as center}, which uses an object's topmost point (instead of center) to trace its trajectory as the topmost position is less likely to be occluded.

\subsection{Simple Physics Simulator (SPS)}\label{ssec:asp-sim}

To better detect collisions, we introduce a Simple Physics Simulator (SPS) with two functions. The first is to predict the linear trajectory of each object after some frame. The second is collision detection.  


\smallskip
\noindent{\bf Linear Trajectory Predictions~~} 
Using several kinematic equations, the simulated object trajectory is computed. Estimations for the coefficient of friction and direction of motion for each object are computed from the perception results from ${\cal M}_p$ with or without IOD. 
This information is later used in equations to calculate the linear distance traveled and positions of each object $o$ in each frame after $o$ is removed from $\mathbb{O}_p$ (i.e., the set of objects whose perception states can be used at the moment) in step 6 of Algorithm~\ref{alg:enhanced_sim}.

From the visual perception module in Figure~\ref{fig:our_model},
we know $x$ and $y$ coordinates of objects in each video frame. We denote the position of an object to have the form of a vector $\mathbf{x}=x\hat{\boldsymbol{\imath}} + y\hat{\boldsymbol{\jmath}}$. $\hat{\boldsymbol{\imath}} $ and  $ \hat{\boldsymbol{\jmath}}$ are unit vectors representing the \textit{x} and \textit{y} direction. We can compute the approximate velocity of an object in some frame up to the end of the video as:
$$\mathbf{v}_{i+t}=(x_{i+t}-x_i)\hat{\boldsymbol{\imath}} + (y_{i+t}-y_i)\hat{\boldsymbol{\jmath}}$$
where 
$i+t\leq max_v$; 
$max_v$ is the last frame in the video and is 127 as CLEVRER videos contain 127 frames;
$\mathbf{v}_i$ represents the approximate velocity of this object in frame $i$; 
$t$ is the temporal resolution, that is, the number of frames between every two consecutive positional information for $x$ (or $y$). In general, the higher the temporal resolution, the less accurate the simulation. 
For NS-DR, since the position of each object is available every 5 frames in the perception results, we set $t=5$. When using IOD or VRDP, $t=1$, since positional information is available for every frame.

The magnitude of the velocity of an object with friction over time is modeled with the following equation:

$$|\mathbf{v}_{i+t}|=|\mathbf{v}_{i}| - g\sigma t$$

\begin{figure}[h]
\centerline{\includegraphics[width=0.5\linewidth]{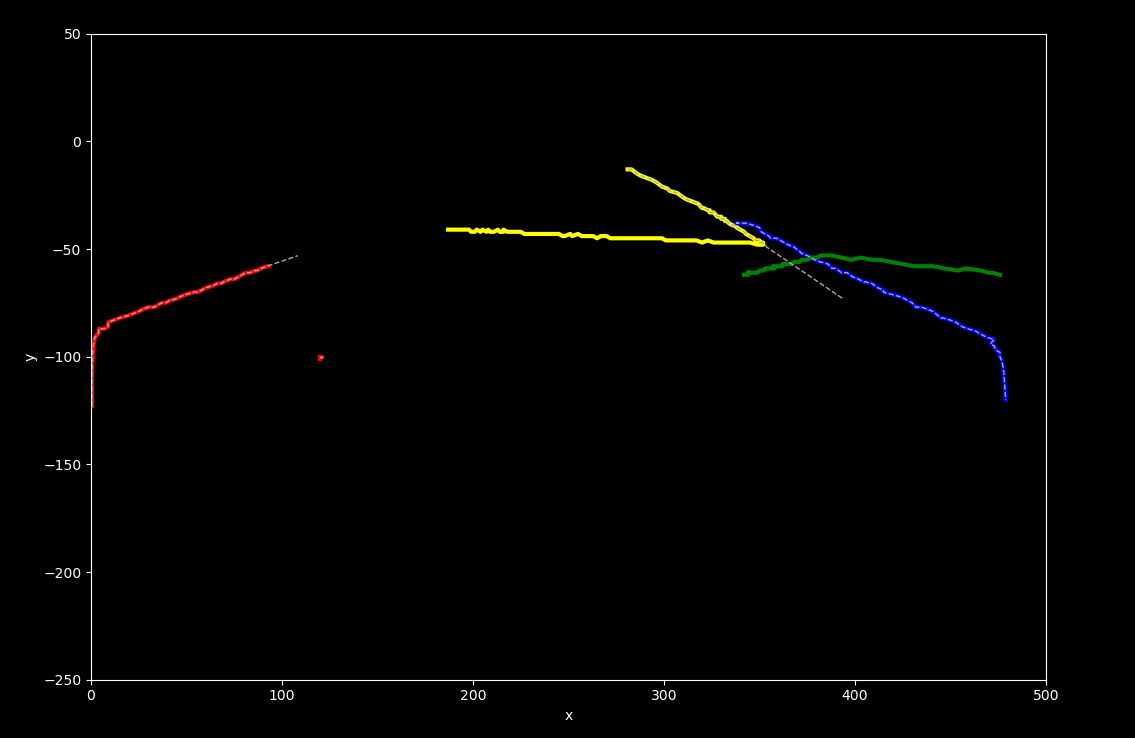}}
\caption{Simple Physics Simulator (SPS) output for the question ``Without the green cylinder, what will happen?'' The colored lines are in-video trajectories detected by the perception model ${\cal M}_p$ with IOD. The gray dashed lines are the full counterfactual trajectories computed by SPS without the green object. The yellow object hits the green object in the video, but SPS simulates the yellow object's physics trajectory from frame 46, just before this collision would happen, according to the {\tt sim(0,46)} fact present in the answer set.} 
\label{phys_module_example}
\end{figure}

\noindent{where $|\cdot |$ denotes the vector norm, $\sigma$ is the coefficient of friction, and $g$ is the gravitational constant. 
%
%
We use the last two perception data (i.e., the two perception states for each object at frames $i$ and $i+t$) and solve for $g\sigma$ for each object.
}
With the frictional term $g\sigma$ and the velocity $v_i$ of each object, 
we can approximate the magnitude of velocity $|\mathbf{v}_{i+t}|$
at frames before the object comes to a stop. 
Furthermore, we can approximate the distance traveled by an object in $t$ frames with:
$$\mathbf{\Delta{x_i}}=|\mathbf{v_i}|\cdot t$$
Finally, we get the simulated trajectory of all objects by splitting up the distance traveled into the $x$ and $y$ components. 
Figure~\ref{phys_module_example} shows an example simulated trajectory in the 2D space.
Note that all objects need to be simulated either from the frame identified by a sim node (e.g., the yellow object in Figure~\ref{phys_module_example}) or from frame $max_v$ (e.g., the red object in Figure~\ref{phys_module_example}).

\smallskip
\noindent{\bf Collision Detection~~} 
To detect a collision,  we check the distance of each object relative to every other object at each time frame. If they are within some threshold, then we detect a collision. The threshold we use is $23.0$ units, learned from the validation set. 

\subsection{Achieving SOTA for the CLEVRER Challenge}
The IOD module serves as a post-processor of a perception model to improve its perception accuracy, thus is applicable to all question types. The SPS module is only applicable to predictive and counterfactual questions.

\medskip
\noindent
{\bf Descriptive and Explanatory Question Answering}\ \ \   
%

Table~\ref{tab:descriptive} shows the improvements due to IOD on the 54,990 descriptive questions in the validation set. 
NS-DR achieves 88.02\% accuracy for descriptive questions, and with IOD,
which makes the prediction to adhere to physical constraints, the accuracy is improved to 95.46\%. VRDP's 93.79\% accuracy is improved to 96.64\% with IOD.

\begin{table}[h!]
\caption{Ablation study on descriptive questions in the validation set. (IOD: improved object detection)}
\centering
\begin{tabular}{@{}lcccccc@{}}
\toprule
\multirow{1}{*}{Model}  & \multirow{1}{*}{Per Ques.(\%)} \\  \midrule
NS-DR   &	               88.02      \\
VRDP  & 93.79 \\ 
NS-DR + IOD   & 95.46    \\ 

VRDP + IOD  & \textbf{96.64}   \\ \bottomrule
\end{tabular}
\label{tab:descriptive} 
\end{table}

Explanatory questions ask about the cause of an event (collision or object exiting) that happens in the video, so like descriptive questions, they do not require physics simulation. We apply the same enhancement IOD. Table~\ref{tab:explanatory} shows the result of each enhancement on the 8,488 explanatory questions and 30,697 options in the validation set. Overall, we achieve 99.68\% question accuracy over baseline NS-DR's 79.31\%. 

\begin{table}
\caption{Ablation study on explanatory questions in the validation set.  (IOD: improved object detection)}
\centering
{
\begin{tabular}{@{}lcccccc@{}}
\toprule
\multirow{1}{*}{Model}  & \multirow{1}{*}{Per Opt.(\%)} & \multirow{1}{*}{Per Ques.(\%)} \\ \midrule
NS-DR            	 & 87.19    		 & 79.31     \\
NS-DR + IOD    	 & \textbf{99.90}  		     & \textbf{99.68}     \\
VRDP         		 & 92.98   			 & 89.00  \\
VRDP + IOD   		 & 99.28 			 & 98.82  \\ \bottomrule
\end{tabular}
}
\label{tab:explanatory}
\end{table}

The improvements we made for descriptive and explanatory questions are relatively simple. However, it is also worth noting that such simple improvements achieve near 100\% accuracy. Such is not the case with end-to-end models.

\medskip
\noindent
{\bf Predictive Query Answering}\ \ \  
%
%
Recall that	predictive questions are about collision events after the video ends. When NS-DR evaluates predictive questions, the symbolic executor calls the functional module \textit{unseen\_events}, which  returns the post-video collision events that PropNet generates. It then checks if any of these predicted collision events match one of the collision options in the question. 
Like counterfactual QA, it is essential to use a simulation to predict the movement, but we observe that the PropNet prediction is the primary source of errors in the validation set, and the errors are overwhelmingly false negatives ($\approx 89.06\%$). 
We find that VRDP also has a high false negative rate of 91.8\% (269 out of 293 errors in the validation set) though the errors are significantly less than NS-DR.


To alleviate the false negative issue, 
we use the simple physics simulator (SPS) (Section~\ref{ssec:asp-sim}) that computes objects' linear trajectories and collision events using physics equations.
%
%
%
The inputs to SPS are the trajectories and collision events generated from the IOD module (Section~\ref{ssec:iod}). These additional collision events found by SPS are appended to the set of post-video collision events predicted in the baseline. 
Finally, the answer is computed by the program executor with the functional programs generated from the baseline question parser.


\begin{table}
\caption{ 
Ablation study on predictive questions in the validation set. (IOD:  Improved object detection, SPS:  Simple Physics Simulator)}
\centering
{
\begin{tabular}{@{}lcccccc@{}}
\toprule
\multirow{1}{*}{Model} &  \multirow{1}{*}{Per Opt.(\%)} & \multirow{1}{*}{Per Ques.(\%)} \\ \midrule
NS-DR &  83.68   & 70.03      \\
NS-DR + SPS &  86.91   & 76.61     \\
NS-DR + IOD + SPS &  89.50   & 79.34     \\
VRDP  &  95.88   & 91.90      \\
VRDP + SPS &  \textbf{96.43}   & \textbf{92.94}     \\
VRDP + IOD + SPS &  95.83   & 91.82     \\
\bottomrule
\end{tabular}
}
\label{tab:predictive_vrdp}
\end{table}

Table~\ref{tab:predictive_vrdp} shows the results of applying IOD and/or SPS on the baselines NS-DR and VRDP for the 3,557 predictive questions and 7,114 options in the validation set. Starting from NS-DR, our best enhancements yield an additional 9.31\% question accuracy and, starting from VRDP, an additional 1.04\% question accuracy.
Adding only the SPS module to the baseline shows a noticeable improvement because the module could find a significant number of missing predictions.

\medskip
\noindent
{\bf Counterfactual Query Answering}\ \ \  
NS-DR and VRDP address counterfactual questions by the simulation to predict collision events when some object is removed. For the simulator, NS-DR uses PropNet and VRDP uses an impulse-based differentiable rigid-body simulator. As with predictive questions, the physics simulation often makes mistakes. 
For NS-DR baseline, out of the total 7,337 errors in the validation options, we find that 3,050 (41.57\%) of them are false positives (i.e., the collision in the option is incorrectly predicted in the set of collision events produced by PropNet) and 4,287 (58.43\%) are false negatives (i.e., the collision in the choice should be predicted by PropNet but not).
VRDP performs much better, with 1756 total errors, where 49.1\% are false positives and 50.9\% are false negatives.

Table~\ref{tab:counterfactual} shows an ablation study on different modules, including our main method $\ours$ and two simple modules IOD and SPS,
on the 9,333 counterfactual questions and 33,051 choices in the validation set.

For NS-DR, we applied $\ours^{approx}$ to directly enhance the predictions of NS-DR, which we denote by $\ours^{approx}_{NSDR}$. We also applied IOD and SPS to improve the accuracy of the perception states and the simulated states. Table~\ref{tab:counterfactual} shows that each enhancement could improve the baseline NS-DR's performance and the best accuracy, 91.49\% per option and 75.39\% per question, is achieved when all enhancements are applied. Consider the best combination, i.e., row (e) in Table~\ref{tab:counterfactual}, where we use $\ours^{approx}_{NSDR}$ with IOD and SPS. 
Among 33,051 question-option pairs in the validation dataset, the results for 20,234 are {\tt determined} by $\ours^{approx}$ with an accuracy of 97.07\%. For the remaining 12,817 question-option pairs, the prediction is the same as NS-DR enhanced with IOD and SPS and the accuracy is 82.75\%.

For VRDP, we applied CRCG using the perception model and the simulation model in VRDP, which we denote by $\ours_{VRDP}$. For comparison, we also applied $\ours^{approx}$ to directly enhance the predictions of VRDP, which we denote by $\ours^{approx}_{VRDP}$. Table~\ref{tab:counterfactual} shows that, while $\ours^{approx}$ could improve the accuracy of the predictions from VRDP, the accuracy can be further improved with $\ours$, which models the influence from other simulated objects with the frame-by-frame simulation in Algorithm~\ref{alg:enhanced_sim}. 
Consider row (h) in Table~\ref{tab:counterfactual} where $\ours^{approx}$ is applied to VRDP. Among 33,051 question-option pairs in the validation dataset, the results for 20,776 are {\tt determined} by $\ours^{approx}$ with an accuracy of 97.38\%. For the remaining 12,275 question-option pairs, the prediction is the same as the baseline VRDP and the accuracy is 94.10\%. (The number of {\tt determined} question-option pairs is slightly different for NS-DR and VRDP because they use different perception model ${\cal M}_p$ to detect objects $\mathbb{O}$ and in-video collisions $\mathbb{C}$.)

%
%


%
Note that when the answer is determined the accuracy is significantly better than when it is not because perception result is being used and is more reliable. This is evidenced by (a) vs. (c); (b) vs. (d); and (f) vs. (h).
For the best results with each respective baseline, even when the answer is not determined, we do not blindly apply simulation, which yields the performance gain evidenced by (d) vs. (e) with the use of SPS and (h) vs. (i) with the use of Algorithm~\ref{alg:enhanced_sim} in $\ours_{VRDP}$ . 
The result justifies the model's prioritization of using perception results rather than the more error-prone baseline counterfactual simulation.

\begin{table}
\caption{Ablation study on counterfactual questions in the validation set. The components are (1) NSDR, (2) VRDP, (3) causal reasoning with ASP ($\ours$), (4) Improved object detection (IOD), and (5) the simple physics module (SPS)} 
\centering
{
\begin{tabular}{@{}lcccccc@{}} 
\toprule
\multirow{1}{*}{Model} &  \multirow{1}{*}{Opt.(\%)} & \multirow{1}{*}{Ques.(\%)} \\ \midrule  
(a)~NS-DR       &   73.98   &   41.55   \\
(b)~NS-DR + IOD       &   75.99   &   43.05   \\
(c)~$\ours^{approx}_{NSDR}$     &   86.22   &   64.59   \\
(d)~$\ours^{approx}_{NSDR}$ + IOD     &   88.95   &   70.78   \\
(e)~$\ours^{approx}_{NSDR}$ + IOD+ SPS      &   91.49   &   75.39   \\ 
(f)~VRDP        &   94.69   &   84.11   \\
(h)~$\ours^{approx}_{VRDP}$        &   95.80   &   87.21   \\ 
(i)~$\ours_{VRDP}$      &   \textbf{96.16}   &   \textbf{88.13}  \\
\bottomrule
\end{tabular}
}
\label{tab:counterfactual}
\end{table}

\section{ASP Programs}\label{appendix:sec:asp}

\subsection{ASP Input Generation}\label{appendix:subsec:asp_input_gen}

The first step in $\ours$ in Figure~\ref{fig:our_model} is to turn (i) questions and options extracted by the Question Parser and (ii) object features and collision events extracted by the Visual Perception module into ASP facts. 

\begin{figure} [ht!]
\centerline{\includegraphics[width=0.8\linewidth]{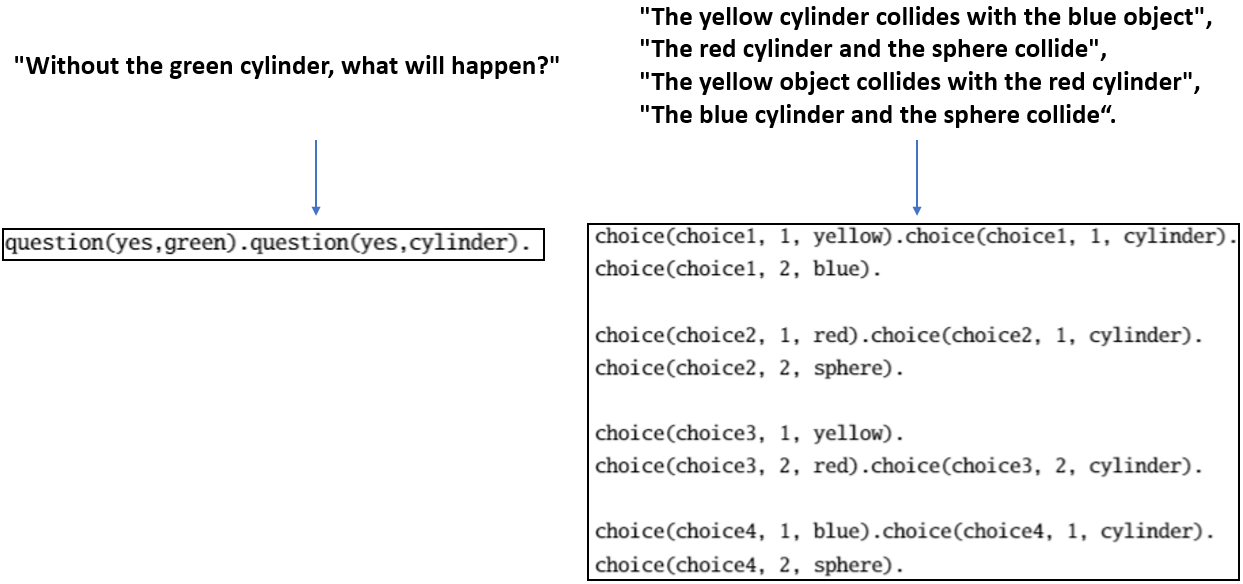}}
\caption{
Conversion of a question and choices into ASP facts.
}
\label{fig:NL_to_ASP}
\end{figure}

For instance, for the question ``Without the green cylinder, what will happen?'' and its four options in the CLEVRER dataset,
Figure~\ref{fig:NL_to_ASP} shows the conversion into ASP facts, which is done by a Python script. 

\begin{figure} [ht!]
\centerline{\includegraphics[width=0.5\linewidth]{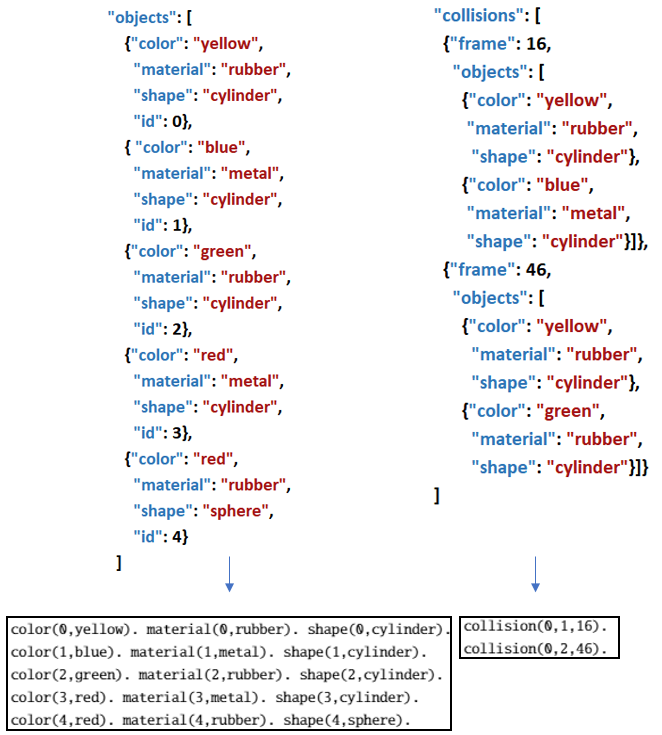}}
\caption{
Conversion of improved object detection (video frame parser) results into ASP facts.
}
\label{fig:traj_to_ASP}
\end{figure}

Figure~\ref{fig:traj_to_ASP} shows the conversion from IOD improved object detection results (from the visual perception module) into ASP facts. These ASP facts along with any post-video collisions detected by the simulation module (without removing any objects) make up the \texttt{input.lp} file in Appendix~\ref{appendix:subsec:asp:crcg}. The post-video collisions as input are necessary because the counterfactual questions in the CLEVRER dataset ask about hypothetical events that may happen during the counterfactual simulation for the duration of the video \textit{and} some time after (we simulate/reason about up to 185 frames, 58 more than the video has).

\subsection{The Full ASP Program for $\ours$}\label{appendix:subsec:asp:crcg}
Below we show the full ASP programs, including ``\texttt{input.lp}'' and ``\texttt{causal.lp}'', that realize our methods in Sections 3.2 and 3.3. The ASP program ``\texttt{input.lp}'' is constructed from video \#147, question \#15 in the validation set of CLEVRER dataset. The ASP program ``\texttt{causal.lp}'' is general and is applied to all examples.




\subsubsection{\texttt{input.lp}}
\texttt{input.lp} encodes the information from a question, its 4 choices, and the corresponding video.
\begin{lstlisting}
color(0,yellow). material(0,rubber). shape(0,cylinder).
color(1,blue). material(1,metal). shape(1,cylinder).
color(2,green). material(2,rubber). shape(2,cylinder).
color(3,red). material(3,metal). shape(3,cylinder).
color(4,red). material(4,rubber). shape(4,sphere).

collision(0,1,16). 
collision(0,2,46). 
collision(3,0,155). 

question(yes,green).question(yes,cylinder).

choice(1, 1, yellow).choice(1, 1, cylinder).
choice(1, 2, blue).

choice(2, 1, red).choice(2, 1, cylinder).
choice(2, 2, sphere).

choice(3, 1, yellow).
choice(3, 2, red).choice(3, 2, cylinder).

choice(4, 1, blue).choice(4, 1, cylinder).
choice(4, 2, sphere).
\end{lstlisting}

\subsubsection{\texttt{causal.lp}}
``\texttt{causal.lp}'' encodes general knowledge about the causal graph and related definitions. 
\begin{lstlisting}
%%%%%%%%%%%%%%%%%%%%%%%%%%%%%%%%
% Rules as interface to turn the atoms in input.lp into more general forms
%   * New atoms:
%      counterfact(remove, qobj(I))
%      option(OptionIdx, qobj(I1), Event, qobj(I2))
%      feature(qobj(I), Feature)
%      query(negated) --- which represents "the question is asking about something not happening"
%%%%%%%%%%%%%%%%%%%%%%%%%%%%%%%%

counterfact(remove, qobj(0)).

option(C, qobj(C*10 + 1), collide, qobj(C*10 + 2)) :- choice(C,_,_).

feature(qobj(0), Feature) :- question(_,Feature).
feature(qobj(C*10 + I), Feature) :- choice(C, I, Feature).

query(negated) :- question(no,_).

%%%%%%%%%%%%%%%%%%%%%%%%%%%%%%%%
% Supress warnings of "atom does not occur in any rule head"
%%%%%%%%%%%%%%%%%%%%%%%%%%%%%%%%

#defined size/2.
#defined enter/3.
#defined query/3.

%%%%%%%%%%%%%%%%%%%%%%%%%%%%%%%%
% Helper atoms
%   * turn size/2, color/2, shape/2 into feature/3 and feature/2
%   * immovable/1 denotes "background" objects that will never move
%   * event/4 denotes the events in {collide, enter}
%   * pos_result/1 denotes the possible result in {yes, no}
%%%%%%%%%%%%%%%%%%%%%%%%%%%%%%%%

feature(O, size, V) :- size(O, V).
feature(O, color, V) :- color(O, V).
feature(O, shape, V) :- shape(O, V).
feature(O, material, V) :- material(O, V).

feature(O, V) :- feature(O, _, V).

immovable(O) :- feature(O, shape, basket).
immovable(O) :- feature(O, shape, ground).

event(O1, collide, O2, Frame) :- collision(O1, O2, Frame).
event(O1, enter, O2, Frame) :- enter(O1, O2, Frame).

pos_result(yes; no).

%%%%%%%%%%%%%%%%%%%%%%%%%%%%%%%%
% Rules for the causal graph
%   * same/2 identify the objects in query with the objects in video
%   * removed/1 denotes the removed object(s)
%   * timestamp/1 denotes the frames with event happening
%   * ancestor/4 determines the ancestor relationships between 2 collisions
%   * affected/2 denotes which nodes in the graph are affected by the removed object
%   * sim/2 represents sim node, which denotes when to start simulation for an object
%%%%%%%%%%%%%%%%%%%%%%%%%%%%%%%%

% Each object in query should be the same as an object in video

same(qobj(I), O) :- feature(O,_,_), feature(qobj(I),_), feature(O,_,T): feature(qobj(I),T).

% Define removed object(s)

removed(O) :- counterfact(remove, qobj(I)), same(qobj(I), O).

% Find all timestamps to be considered in the causal graph

timestamp(T) :- collision(_,_,T).
timestamp(T) :- enter(_,_,T).

% Collision is symmetric

collision(O2,O1,T) :- collision(O1,O2,T).

% The ancestor relation is introduced either by the same object with different frames, or by a collision, or 

ancestor(O,T1,O,T2) :- feature(O,_,_), timestamp(T1), timestamp(T2), T1<T2, not immovable(O).
ancestor(O1,T,O2,T) :- collision(O1,O2,T), not immovable(O1), not immovable(O2).
ancestor(O1,T1,O2,T2) :- ancestor(O1,T1,O3,T3), ancestor(O3,T3,O2,T2), (O1,T1)!=(O2,T2).

% Find the nodes (i.e., object states) in the graph that are affected

affected(O,T) :- removed(O), collision(_,_,T).
affected(O,T) :- removed(O'), ancestor(O',T',O,T).
% If we can remove "anything", every node that has an ancestor is affected
affected(O,T) :- counterfact(remove, any), ancestor(O',_,O,T), O!=O'.

% Find the sim nodes in the graph

sim(O,T) :- not removed(O), affected(O,T), T<=T': affected(O,T').

% The result is determined to be yes if the collision happened in the video at a time when the states of o1 and o2 are not affected by the removed object.

determined(Q, yes) :- option(Q, qobj(I1), Event, qobj(I2)), 
    same(qobj(I1), O1), same(qobj(I2),O2),
    event(O1, Event, O2, F),
    not affected(O1,F), not affected(O2,F).

% The result is determined to be no if O1 or O2 is removed.

determined(Q, no) :- option(Q, qobj(I1), Event, qobj(I2)), 
    same(qobj(I1), O1), same(qobj(I2),O2),
    removed(O1): not removed(O2).

% The result is determined to be no if the collision did not happen in the video and the states of O1 and O2 are not affected by the removed objects.

determined(Q, no) :- option(Q, qobj(I1), Event, qobj(I2)), 
    same(qobj(I1), O1), same(qobj(I2),O2),
    not event(O1, Event, O2, _),
    not affected(O1,_), not affected(O2,_).

% we answer negated result if the query is asking about an event not happening

answer(Idx, Ans) :- determined(Idx, Res), 
    pos_result(Res), pos_result(Ans),
    Res = Ans: not query(negated);
    Res != Ans: query(negated).

% we answer the count if the query is about counting events

answer(N) :- query(counting, Event, qobj(I)), 
    same(qobj(I), O),
    N = #count{Ox: event(Ox,Event,O,_), not removed(Ox)},
    not sim(Ox,_): feature(Ox,_,_).

% we answer tbd if no result is predicted

{answer(Idx, tbd)} :- option(Idx,_,_,_).
:- option(Idx,_,_,_), #count{Res: answer(Idx, Res)} = 0.
:- option(Idx,_,_,_), answer(Idx, tbd), #count{Res: answer(Idx, Res)} > 1.

#show sim/2.
#show answer/2.
\end{lstlisting}

\end{document}